\documentclass{article}
\usepackage{spconf,amsmath,graphicx}
\usepackage{algorithm}
\usepackage{algorithmic}
\usepackage{amsfonts}
\usepackage{cleveref}
\usepackage{url}
\usepackage{cite,balance}
\usepackage{textcomp}

\title{SUPERVISED ENCODING FOR DISCRETE REPRESENTATION LEARNING}
\name{Cat P. Le$^{\star}$ \qquad Yi Zhou$^{\dagger}$ \qquad Jie Ding$^{\ddagger}$ \qquad Vahid Tarokh$^{\star} \thanks{This work was supported in part by Office of Naval Research Grant No. N00014-18-1-2244.}$}

			\address{
			$^{\star}$ Department of Electrical and Computer Engineering, Duke University \\ $^{\dagger}$ Department of Electrical and Computer Engineering, University of Utah \\ $^{\ddagger}$ School of Statistics, University of Minnesota}
			
\begin{document}
\maketitle
\begin{abstract}
Classical supervised classification tasks search for a nonlinear mapping that maps each encoded feature directly to a probability mass over the labels. Such a learning framework typically lacks the intuition that encoded features from the same class tend to be similar and thus has little interpretability for the learned features. 
In this paper, we propose a novel supervised learning model named Supervised-Encoding Quantizer (SEQ). The SEQ applies a quantizer to cluster and classify the encoded features. We found that the quantizer provides an interpretable graph where each cluster in the graph represents a class of data samples that have a particular style. We also trained a decoder that can decode convex combinations of the encoded features from similar and different clusters and provide guidance on style transfer between sub-classes. 
\end{abstract}

\section{Introduction}
\label{sec:intro}
Although deep learning has achieved great success in solving a variety of challenging tasks,  most of the existing deep learning models do not provide sufficient interpretability for the learned features. For instance,  classical supervised learning only maps the training samples to their corresponding labels, and often do not exploit the features of the samples. Consequently, such a learning framework typically produces a black-box model that provides little interpretability for the learned classification rule. As modern machine learning tasks involve increased model complexity and dimensionality, there is a rising interest in developing deep learning models that can provide a good visualization of the features and interpretable decision rules. 

In this work, we combine a supervised-encoding technique with the \textit{k}-means quantizer to build a supervised learning model with high classification accuracy and interpretability. Briefly speaking, we first train an encoder network  that maps  input data to low-dimensional embeddings with a soft-max layer and the cross-entropy loss using the standard supervised training. Such a pre-trained encoder ensures the encoded features of the training data samples to be linearly separable with regard to the labels. 
Then, the encoded features of the training data are clustered by a \textit{k}-means quantizer, and each cluster is assigned a label by performing a majority vote. This quantization approach with supervised encoding enables the model to perform regular classification tasks and identify sub-classes of the data that correspond to different styles. Lastly, we also train a decoder that can reconstruct the data from its features and hence can generate new data samples with specified style mixtures.

\textbf{Related Work}.
 Applying quantization to the embedded features has been studied before, e.g. in~\cite{xie2016unsupervised,dilokthanakul2016deep,pesteie2018DeepNM,agustsson2017soft}, but it remains a challenge to achieve desired learning performance. In particular, some authors have suggested using a non-linear mapping to pre-process the data before performing the quantization, e.g. using an autoencoder for dimension reduction  \cite{xie2016unsupervised,aytekin2018clustering,pesteie2018DeepNM}. In \cite{xie2016unsupervised}, DEC model introduces an incorporated loss to jointly train an autoencoder and a quantizer. As a result, it improves the separation of data and makes the quantization process more efficient. The CAE-$l_2$ is another model proposed in \cite{aytekin2018clustering} that consists of a convolutional autoencoder and a \textit{k}-means quantizer. The model makes data features more separable in the Euclidean space by applying a $l_2$ normalization to the encoded features. Recently, \cite{pesteie2018DeepNM} developed a DNM model that uses a self-organizing map \cite{kohonen1990self, kohonen1996engineering} as the quantizer to map the embedded data into a predefined number of clusters. This approach can learn multiple different styles of the data within the same class, but it cannot perform classification tasks.
 
 Our method is related to some prior work in image compression using deep neural networks. For example, a quantization method using soft assignments over time was proposed to obtain a hard clustering \cite{agustsson2017soft}. Compressive autoencoder \cite{theis2017lossy} compresses lossy features using rounding-based quantization before entropy encoding. Our approach also connects to the literature of learning the distribution of data representations. In \cite{dilokthanakul2016deep}, GMVAE uses Gaussian mixture model on the latent space of VAE \cite{kingma2013auto} to understand the distribution of latent representations. VQ-VAE is a model proposed in \cite{van2017neural} that uses VQ as a bridge to connect the encoder and the decoder of VAE. These approaches learn the discrete latent distribution and can be considered as generative models.

\section{Supervised-encoding Quantizer}
\label{sec:format}
\begin{figure}[t]
	\centering
	\centerline{\includegraphics[width=8.5cm]{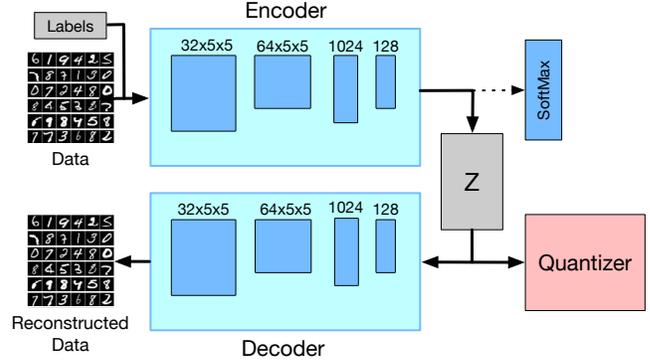}}
	\caption{Illustration of the proposed supervised-encoding quantizer (SEQ).}\label{fig: 1}
\end{figure}

Our supervised-encoding quantizer (SEQ) model consists of an encoder, a quantizer, and a decoder, as illustrated in \Cref{fig: 1}. 
The SEQ training consists of the following steps, which are further elaborated in the following subsections.
\begin{enumerate}
	\item Encoding: we pre-train the encoder by attaching a soft-max layer to its output and train the encoder via standard supervised training;
	\item Quantization: the encoded features produced by the pre-trained encoder are passed to the quantizer for clustering, as illustrated in \Cref{fig: 2};
	\item Decoding: we further train a decoder that can reconstruct the original data samples from the encoded features.    
\end{enumerate}
Regarding the quantization step, various quantizers can be applied to the SEQ model, including \textit{k}-means, vector quantization, self-organizing map, grow-when-required network \cite{marsland2002self}, and Gaussian mixture model. For simplicity, we apply the \textit{k}-means clustering algorithm \cite{forgy1965cluster} in the quantization step. We elaborate on each of the main steps as follows.

\subsection{Pre-training Encoder}
\label{ssec:train-encoder}
To obtain an interpretable embedding space, the encoder of the SEQ needs to be pre-trained with labeled data. 
First, we construct an encoder consisting of linear layers and/or convolutional layers. To train the encoder, we attach a Softmax layer to the output of the encoder. Then, we train such a neural network using the labeled data with cross-entropy loss and stochastic gradient descent. After the training, we remove the Softmax layer and keep the parameters of the encoder. In particular, the output embedding features $Z$ produced by the pre-trained encoder are guaranteed to be linearly separable due to the supervised training with cross-entropy loss. 

\subsection{\textit{k}-means Quantization}
\label{ssec:train-quantizer}
\begin{figure}[t]
	\centering
	\centerline{\includegraphics[width=8.5cm]{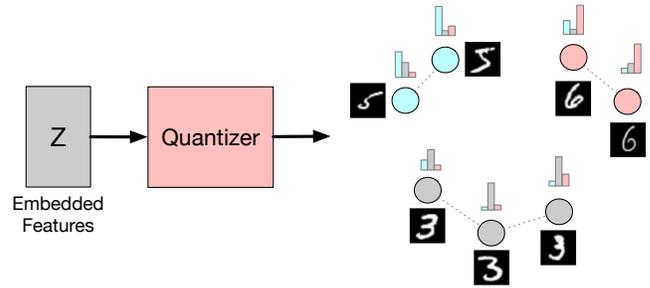}}
	\caption{Illustration of the quantization process.}\label{fig: 2}
\end{figure}

The pre-trained encoder maps each data sample $x$ to a corresponding embedded feature $Z(x)$. Then, as illustrated in \Cref{fig: 2}, the quantizer takes the embedded features $Z$ of all data samples as the input and applies the \textit{k}-means clustering algorithm to quantize these features with a pre-defined number $K$. The clustering produces $K$ clusters with centers $C_1, C_2, ..., C_K$, and each of the features is assigned to the cluster that has the minimum distance between the feature and the cluster center. Such a clustering result naturally produces a topological graph, where each cluster consists of the features of samples that are close to each other in terms of the Euclidean distance. By choosing $K$ to be greater than the total number of classes, the \textit{k}-means clustering can identify the sub-classes of the samples within each class of the data. To further perform a classification task, we use a simple approach based on the majority vote strategy, called histogram-based labeling. In particular, the label for each cluster is set to be the label with the highest frequency in that cluster.

In practice, we found that a larger $K$ usually leads to better classification performance but more computational costs. To tune an appropriate choice of the total number of clusters $K$ for the SEQ quantizer, we propose the following simple rule. We denote $P_Q$ as the classification accuracy of the quantizer, and $P_{E}$ as the classification accuracy of the encoder network achieved in the pre-training phase. Since the accuracy of the quantizer is upper bounded by that of the encoder, we search for the smallest value $K$ such that $P_Q > P_E - \epsilon$.

\subsection{Training Decoder}
\label{ssec:train-ae}
We attach a decoder to the output of the encoder in order to reconstruct the data from the embedded features. First, we construct a decoder whose layers are symmetrical to the encoder's structure. To train the decoder, we fix all the parameters of the encoder and apply the MSE loss to measure the reconstruction error on the training data samples as:
\[
\mathcal{L}(\theta) := 
\Vert  X - \mathcal{D}_\theta(\textbf{sg}[\mathcal{E}(X)]) \Vert^2 ,
\]
where $\mathcal{E}$ denotes the pre-trained encoder that compresses the data samples $X$ into embedded features $Z$, \textbf{sg} denotes the stop-gradient operator which freezes the encoder's parameters, $\mathcal{D}_\theta$ denotes the decoder with the parameters $\theta$. We minimize the reconstruction loss $\mathcal{L}$ over the decoder parameters $\theta$ \textit{only} using the stochastic gradient descent algorithm.

\section{Experimental Study}
\label{sec:pagestyle}

\subsection{Performance Evaluation}
We evaluate the classification accuracy of SEQ on both the training dataset and the test dataset. In specific, we consider the following three network structures for the encoder network of the SEQ \footnote{Source code for the SEQ model is available at the GitHub link below: https://github.com/lephuoccat/Supervised-Encoding-Quantizer}:
\begin{enumerate}
	\item LAE-2: the encoder has 2 fully connected hidden layers with dimensions dense(1024)-dense(128);
	\item LAE-4: the encoder has 4 fully connected hidden layers with dimensions dense(1024)-dense(512)-dense(256)-dense(128);
	\item CAE-4: the encoder has 2 convolution layers and 2 fully connected hidden layers  with dimensions $(32\times5\times5)$-$(64\times5\times5)$-dense(1024)-dense(128).    
\end{enumerate}
In addition, regarding the \textit{k}-means quantizer, we consider the total number of clusters $K$ ranging from 10 to 120. We evaluate the performance of these SEQ models on the standard MNIST \cite{lecun2010mnist} and fashion-MNIST \cite{xiao2017fashion} datasets.

We first evaluate the clustering performance of the \textit{k}-means quantizer of SEQ on the training dataset using unsupervised learning criteria. The metric for evaluating the clustering performance is defined by the percentage of training samples that are assigned to the correct label by the \textit{k}-means quantizer as described in \Cref{ssec:train-quantizer}.
We compare the clustering performance of SEQ with those of DEC \cite{xie2016unsupervised}, IDEC \cite{guo2017improved}, DCEC \cite{guo2017deep}, and CAE-$l_2$ \cite{aytekin2018clustering}, respectively. The results are shown in \Cref{tab: 1}. The reported value in the parenthesis is the standard error of our average accuracy from   re-samplings. Standard errors of other methods were not reported in previous works. The table shows that our SEQ achieves significantly higher accuracy on the training dataset compared with state-of-the-art methods.
A possible reason is the advantage brought by the embedded space of SEQ that is well clustered in line with the true labels.

\begin{figure}[t]
  \centering
  \centerline{\includegraphics[width=9cm]{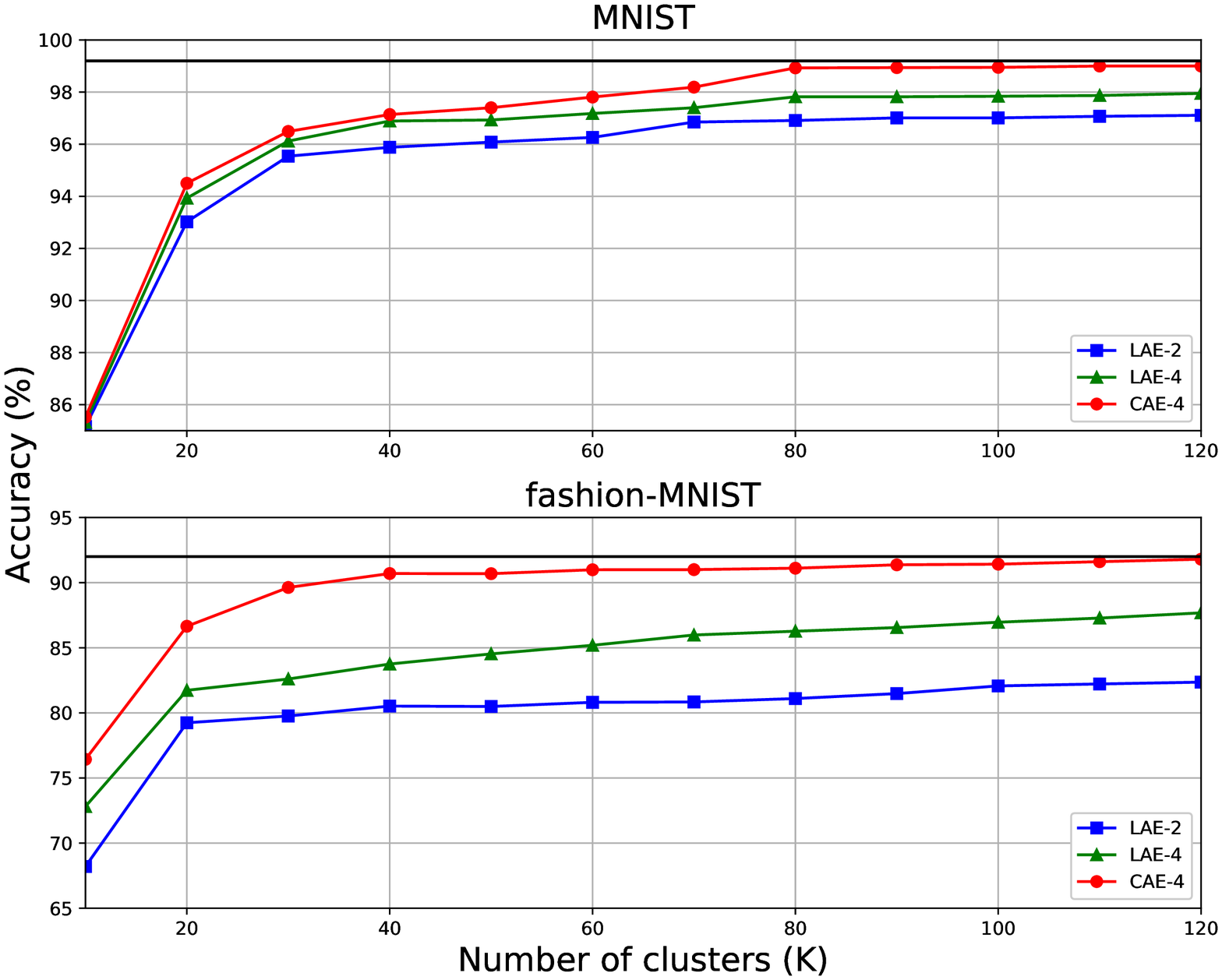}}
  \caption{The classification performance of SEQ with respect to the number of clusters for MNIST and fashion-MNIST.}\label{fig: 3}
\end{figure}

Next, we evaluate the classification performance of SEQ on the test dataset. In specific, for a given new test data sample, we pass it through the encoder to obtain its feature. Then, we identify the cluster with the closest center point to this feature vector and assign the corresponding cluster label to be the predicted label for the test data sample.   
\Cref{fig: 3} shows the classification performance of SEQ on test data versus the number of clusters. We observe a notable improvement of the test accuracy as the number of clusters increases from 10 to 120. 
In particular, the maximum accuracy reaches 99\% as $k \geq 100$ on the MNIST dataset. Also, the test performance improves when the number of layers in the encoder increases. Moreover, the SEQ model that adopts convolutional layers in the encoder outperforms the other models with linear layers in the encoder. Similar observations are made on the fashion-MNIST dataset and the maximum accuracy is 91.8\% when $k \geq 120$. The performance gap of the three SEQ models is more noticeable for  fashion-MNIST, which is considered to be more challenging than  MNIST.
Also, the solid black horizontal line denotes the test accuracy of the pre-trained encoder, which upper-bounds the test performance of SEQ.

\begin{table}[t]
  \centering
  \caption{\label{tab:unspervised-table}The clustering performance on MNIST}\label{tab: 1}
  \begin{tabular}{ cccccc } 
    \hline
    \hline
    DEC & IDEC & DCEC & CAE-$l_2$ & SEQ \\
      & & & \textit{k}-means & \textit{k}-means \\
    \hline
    86.55 & 88.06 & 88.97 & 95.11 & \textbf{99.74 (0.046)} \\ 
    \hline
    \hline
  \end{tabular}
\end{table}

\subsection{Interpretability}
\begin{figure}[t]
  \centering
  \centerline{\includegraphics[width=8.5cm]{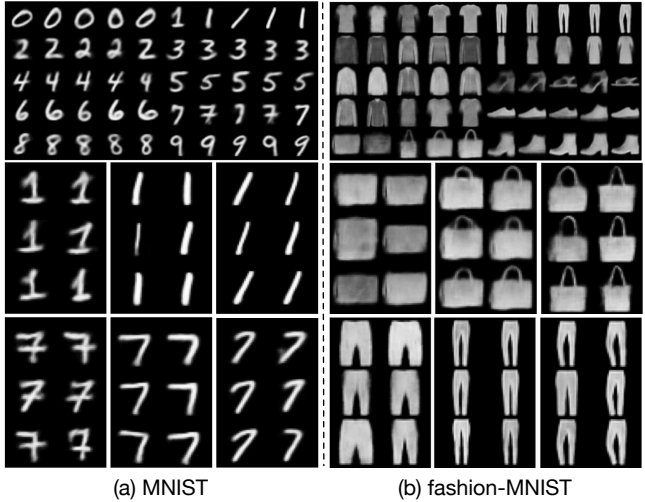}}
  \caption{\textbf{Row 1}: The average decoded images from the clustering points showing multiple styles for each class of data for MNIST and fashion-MNIST. \textbf{Row 2-3}: The mapped data samples from different clusters.}\label{fig: 4}
\end{figure}

 In this subsection, we show that the SEQ model can provide interpretable clustering results. In the top row of \Cref{fig: 4}, we present the images decoded from the average features of 50 cluster centers on the MNIST and fashion-MNIST datasets, respectively. Specifically, each decoded image is obtained by feeding the average of the features within each cluster to the trained decoder. As observed in the top two figures, the average of the features of the clusters in a certain class can be decoded into images with different styles. For example, for the MNIST dataset, the digit 1 and digit 7 have multiple different styles, which are captured by different clusters and are shown in the middle and bottom rows of \Cref{fig: 4} (a). 
 
 Similarly, for the fashion-MNIST dataset, the bag images have three different styles, i.e., \textit{no handle}, \textit{short handle}, and \textit{long handle}, which are presented in the middle row of \Cref{fig: 4} (b). The trousers also have three styles and are shown in the bottom row. The results indicate that the embedding space of the SEQ quantizer is interpretable in the sense that data samples with the same style are clustered together.

\subsection{Semi-generative Model}
Given that the SEQ model is capable of clustering similar data into a sub-classes, we consider the output of this model as a graph network where each node of the graph is the center clustering point.
We also demonstrate that the embedding space of the SEQ model is smooth and continuous in decoding, in the sense that we can combine and interpolate the feature vectors within the same cluster and obtain decoded images with the same style.

\begin{figure}[t]
  \centering
  \centerline{\includegraphics[width=8.5cm]{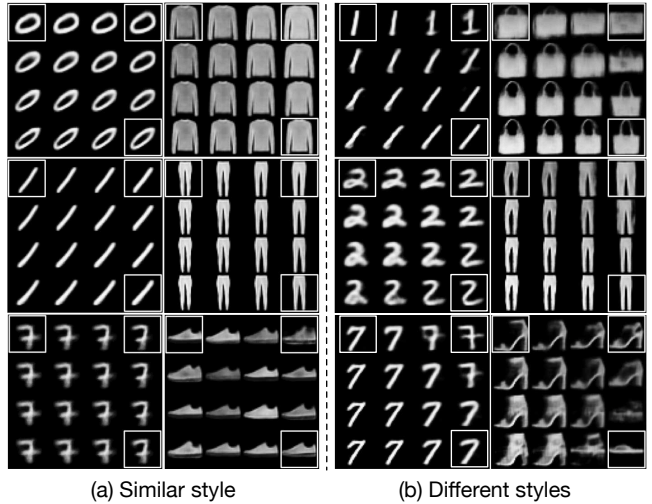}}
  \caption{The generative data using convex hull of three data samples from (a) the same cluster and (b) different clusters for MNIST and fashion-MNIST. The images inside the white boxes are the original data samples.}\label{fig: 5}
\end{figure}

We consider convex combinations of three features within a certain cluster in the form of $x=\sum_{i=1}^{3}\alpha_i x_i$, where we vary $\alpha_1, \alpha_2 \in (0, 0.5)$ and $\alpha_3 = 1-\alpha_1-\alpha_2$. For any such convex combination, the newly generated feature is considered  to belong to the same class. In \Cref{fig: 5} (a), we present the new data samples that are decoded from the features using convex combinations of three features with different combination coefficients. In each block of \Cref{fig: 5} (a), the decoded images of the three selected features (denotes as $x_1, x_2, x_3$) from the same cluster are located at the top left, top right and bottom right, respectively, and are highlighted by the white boxes.  The decoded images from the convex combined features are shown in the blocks. We observe that the generated images have clear shapes and similar structures.

Lastly, we applied convex combinations to three data samples from different clusters (styles) within the same class. As shown in \Cref{fig: 5} (b), the generated images are smooth transitions between different styles. Interestingly, the generated images also have clear shapes and resemble the class structures. Additionally, they do not suffer from any \textit{blurring} artifact, which can be an issue in other generative models such as GANs \cite{goodfellow2014generative} and VQ-VAE \cite{van2017neural}. 

\section{Conclusion}
We provide a new perspective of looking at supervised classification and unsupervised generative models with nonlinear feature mapping.
We first propose a supervised learning model SEQ that applies a quantizer to cluster and classify the encoded features. We surprisingly found that each quantizer provides an interpretable representation of a set of images that have a particular style. As a result, the encoder can be regarded as an interpretable dimension reduction that compresses data in such a manner that closer styles in the data domain correspond to closer features in the feature domain.
From the computation viewpoint, the new supervised learning method requires low-complexity layers for the encoder and decoder which can enable faster training.
We also show how to generate data using the convex hull from a few data samples in the network model. Since the model learns to cluster the data into sub-classes that have different styles, it is able to precisely control the generating process to produce the desired data style.
As an ongoing work, we are applying the developed techniques in  medical applications, e.g. generating specific types of cancer images.

\vfill\pagebreak
\balance
\bibliographystyle{IEEEbib}
\bibliography{refs}

\end{document}